\documentclass[%
 reprint,
 amsmath,amssymb,
 aps,
]{revtex4-2}

\usepackage{graphicx}
\usepackage{dcolumn}
\usepackage{bm}

\begin{document}

\preprint{APS/123-QED}

\title{Iterative Corpus Refinement for Materials Property Prediction\\ Based on Scientific Texts}

\author{Lei Zhang}
\email[Corresponding author: ]{lei.zhang-w2i@rub.de}

\author{Markus Stricker}
\email{markus.stricker@rub.de}

\affiliation{%
Interdisciplinary Centre for Advanced Materials Simulation, Ruhr-University Bochum,~Universitätsstraße 150, 44780 Bochum, Germany
}





\begin{abstract}
The discovery and optimization of materials for specific applications is hampered by the practically infinite number of possible elemental combinations and associated properties, also known as the `combinatorial explosion'. 
By nature of the problem, data are scarce and all possible data sources should be used.
In addition to simulations and experimental results, the latent knowledge in scientific texts is not yet used to its full potential.
We present an iterative framework that refines a given scientific corpus by strategic selection of the most diverse documents, training Word2Vec models, and monitoring the convergence of composition-property correlations in embedding space.
Our approach is applied to predict high-performing materials for oxygen reduction (ORR), hydrogen evolution (HER), and oxygen evolution (OER) reactions for a large number of possible candidate compositions.
Our method successfully predicts the highest performing compositions among a large pool of candidates, validated by experimental measurements of the electrocatalytic performance in the lab.
This work demonstrates and validates the potential of iterative corpus refinement to accelerate materials discovery and optimization, offering a scalable and efficient tool for screening large compositional spaces where reliable data are scarce or non-existent.
\end{abstract}

\maketitle


\section{Introduction}
The discovery of new materials has traditionally relied on experimental intuition and trial-and-error methods, where researchers manually combined and tested materials, guided by experience and theoretical knowledge~\cite{Fujishima20001,Geim2007183,Yeh2004299}.
These methods have led to significant breakthroughs, such as platinum-based electrocatalysts for fuel cells~\cite{Ferreira2005A2256} and advanced alloys for aerospace applications~\cite{Ezugwu2003233}.
However, they are time-consuming, resource-intensive, and difficult to scale, particularly as material systems containing more than one or two principle elements become more complex.
\textit{Modern} material systems, such as high-entropy alloys~\cite{Miracle2017448} or oxides~\cite{Zerdoumi2024}, multi-principal element compounds~\cite{Huang200474}, involve a large number of possible and tunable compositions, which represents a possibility to use them as `discovery platforms'~\cite{Batchelor2019}.
However, the sheer number of possible compositions makes experimental searches impractical or even impossible.
Furthermore, global challenges such as the switch to renewable energy and sustainability require faster and more efficient ways to discover new and optimize existing materials~\cite{Larcher201519,Ragauskas2006484}.

In recent years, computational methods have become powerful tools for materials discovery.
Simulation techniques such as density functional theory (DFT) can predict material properties, reducing the need for extensive experimental trials at the cost of the energy spent in high performance computing centers~\cite{Jain2016,Saal20131501,SCHMIDT2024101560}.
Machine learning has added another dimension by finding patterns in structured datasets and predicting properties for unexplored compositions~\cite{Liu2017159,Raccuglia201673}.
Although these methods have been successful in narrowing the search space, they have limitations.
DFT simulations are computationally expensive, in particular when small $\leq 1\,$atom-\% compositional changes require many calculation to achieve a statistically correct property prediction for one composition point due to the different random distributions of elements and their impact on material properties.
Supervised machine learning models to substitute expensive DFT simulations require high-quality datasets that are available at scale and variety to match the parameter space of technologically interesting materials.

The scientific literature offers an alternative resource for material discovery.
Research articles and patents contain hidden knowledge encoded in text about composition-property relationships from experimental results and theoretical approaches~\cite{Tshitoyan201995}.
Natural language processing (NLP) methods like Word2Vec~\cite{Mikolov2013} and Doc2Vec~\cite{Le20142931} can extract this knowledge by turning textual data into vector representations.
These vectors capture relationships and correlations between words, which allows to link `material dimensions', such as composition, to properties, e.g. electrocatalytic performance, and to develop models for the prediction of new high-performing candidate compositions.
Unlike simulations or structured datasets, text mining can leverage unstructured data, providing access to this latent knowledge.

Despite its potential, using the scientific literature effectively presents challenges.
Not all text sources are equally relevant, and including too many irrelevant or redundant sources can reduce the predictive quality of the models.
The increasing volume of scientific literature, including artificial intelligence (AI)-generated content, makes this issue even more pressing.
Many AI-generated texts are repetitive or inaccurate, introducing noise into the dataset, and if repetetively used to retrain the models eventually leading to their \textit{collapse}~\cite{Shumailov2024,Shumailov2024755}.
Effective methods are therefore needed to filter out `low-quality content' to ensure that models are trained on the most meaningful information.

Another challenge is scalability.
NLP models can process large datasets, but training them on massive corpora is expensive and it is unclear if more data actually results in a better model.
Static, corpus-based models also struggle to adapt to specific tasks.
For instance, a model trained on a broad corpus may not effectively capture the relationships required to predict material properties for a specific system accurately.
The term `material system' here refers to a fixed set of elements which can be mixed in an arbitrary proportion.
To address these challenges, there is a need for methods that allow to tailor a training corpus w.r.t. specific prediction tasks.

In this work, we propose an iterative framework to address these challenges.
First we start with a broad collection of abstracts from scientific papers, furtheron referred to as `documents', and use Doc2Vec embeddings~\cite{Bilgin2017} to create a \textit{map} of abstracts.
This map allows us to then derive an ordered list based on greedy selection which represents the most diverse documents to avoid information duplication and information fuzziness.
From this list, we use batches of 50 documents to train Word2Vec~\cite{Goldberg2014,Mikolov2013} models.
In their respective embedding space, we measure the change of the centroid of the embeddings w.r.t. our specific prediction task and stop adding new document batches once a convergence criterion is met.
The resulting Word2Vec model then constitutes our material system-optimized model which we use to predict reaction-specific high-performance candidate compositions for electrocatalysis.
To demonstrate the effectiveness of our approach, we predict highest performing compositions from three different material systems for three different electrocatalytic reactions: oxygen reduction (ORR), hydrogen evolution (HER), and oxygen evolution (OER) and validate them against experimental measurements.
While electrocatalysts are used as a case study, our framework is general and can be applied to a wide range of materials discovery tasks.

\section{Experiments}

All code and data needed to reproduce our workflow are available. 
The code can be found in~\cite{Author2025}, references to all datasets are provided when they are introduced further below.

\begin{figure}
    \centering
    \includegraphics[width=1\linewidth]{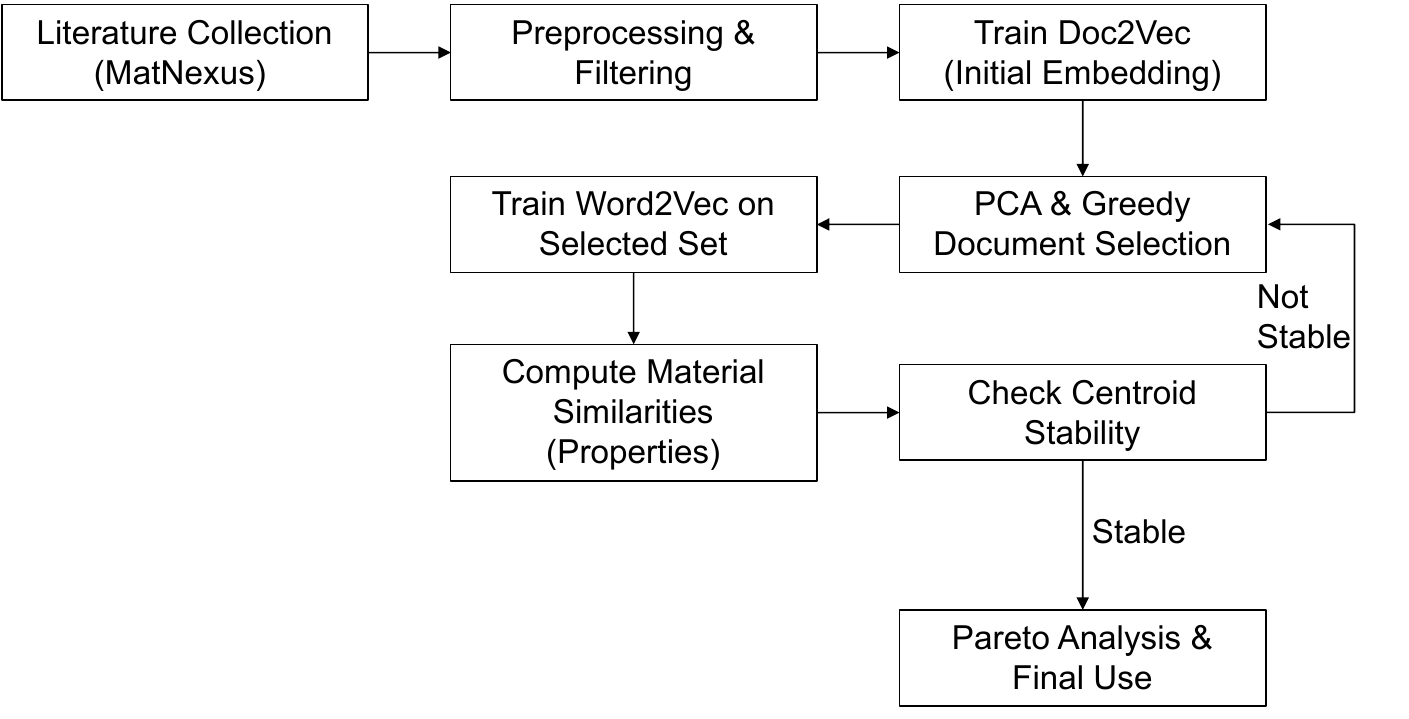}
    \caption{Schematic overview for iterative corpus refinement framework from literature collection, document selection, Word2Vec model training to final predictions based on Pareto front analysis.}
    \label{fig:method_illustration}
\end{figure}

\subsection{Corpus Collection and Preprocessing}

Fig.~\ref{fig:method_illustration} shows a schematic overview of our methodology.
First, we collect a relevant corpus of documents using the \texttt{PaperCollector} module in \texttt{MatNexus}~\cite{Zhang2024}.
As source for the documents we use the application programming interfaces (APIs) of Scopus and and limit ourselves to use open access-only publications up to including the year 2023 which results in 6506 papers.
The retrieved abstracts and metadata are stored in a structured  comma-separated values (CSV) file.
Text preprocessing is performed using the \texttt{TextProcessor} module, which removes licensing statements, filters common English stopwords, and retains domain-specific terms, such as chemical element symbols.
Tokenization is applied to prepare the text for embedding generation.

\subsection{Embedding Generation and Greedy Selection}

The initial document embeddings are generated using a Doc2Vec model trained on the full corpus containing 6506 documents.
The embeddings capture the semantic content of each document, mapping the tokens into a 200-dimensional vector space, which offers a balanced trade-off between computational efficiency and representational capacity for capturing domain-specific linguistic nuances.
A greedy selection algorithm is then applied to iteratively refine the subset of documents for analysis.

\textbf{Initialization:} We initialize the greedy selection algorithm with the central document in embedding space.
This ensures that the starting point is representative of the corpus's overall thematic distribution, i.e. an ``average abstract'' w.r.t. the Doc2Vec embedding space.

\textbf{Iterative Expansion:} Subsequent documents are selected by identifying the farthest (in cosine distance) from all previously selected documents. It is a classic greedy farthest point sampling except the initial point is selected always the central document in embedding space instead of a random starting point.
To improve computational efficiency and focus on principal directions of variance, the 200-dimensional Doc2Vec embeddings are reduced to two dimensions using principal component analysis (PCA) before distance calculations.
The selection process continues until a predetermined batch size of 50 additional documents is reached, which serves as a heuristic balance between introducing sufficient diversity in each iteration and maintaining computational manageability.
This process results in an ordered list of the most different documents in batches of 50 and constitutes the basis for subsequent analysis and optimization.

A selected subset is then used to train a Word2Vec~\cite{Mikolov2013,Goldberg2014} model via the \texttt{Vec\allowbreak-Generator} module of \textsc{MatNexus}~\cite{Zhang2024}.
The skip-gram architecture is used with a vector size of 200, a window size of 5, and hierarchical softmax.
The vector size of 200 provides sufficient capacity to capture nuanced relationships between terms while maintaining training efficiency.
A window size of 5 is used to reflect a moderate contextual range, allowing the model to learn meaningful co-occurrence patterns without overextending semantic connections.
Hierarchical softmax is chosen for its effectiveness in handling large vocabularies with improved training speed over full softmax.

\textbf{Material Similarity Calculation:}
We create the representation of `a material' w.r.t. its composition by a linearly weighted superposition of the embeddings for pure elments, e.g. `Pt' for platinum, 'Pd' for palladium, etc.
The linear weights model the composition.
That is, the representation of a composition of 20\,\% element A and 80\,\% element B means to linearly superpose the word embeddings $R_i$ of elements A and B $R_{A_{20}B_{20}} = 0.2 R_A + 0.8 R_B$.
We then calculate the similarity scores of each composition to the embedding vectors of the two properties `dielectric' and 'conductivity' using cosine similarity which we denote $S_{\text{dielectric}}$ and $S_{\text{conductivity}}$.
Given a composition space and concentration resolution for a given material system containing several elements, we can use the two-dimensional similarity scores to calculate a centroid of the based on $N$ different compositions in the material system as follows:

\begin{equation}
        \text{centroid} = \frac{1}{N}\sum_{i=1}^N 
        \begin{bmatrix}
        S_{\text{dielectric}}(i) \\
        S_{\text{conductivity}}(i)
        \end{bmatrix}.
        \label{eq:centroid}
\end{equation}

\textbf{Convergence Criterion:} 
The previous step is then iterated with an increasing corpus size in batches of 50 documents.
With each increase, we monitor the change of the centroid coordinates using Euclidean distance.
If the distance between the previous and current centroid falls below a heuristically determined threshold of $0.03$, we stop adding more documents and define the word embedding model \textit{converged}.
Our rationale is that when an additional batch of 50 documents does not significantly change the word embeddings, adding more documents is unnecessary or even detrimental to predictive performance.

\subsection{Pareto Optimization for Candidate Selection}

With the converged Word2Vec model for a given material system, Pareto optimization is applied to identify optimal trade-offs between material properties for three electrochemical reactions.
For the Pareto front optimization we follow~\cite{Zhang2025a} which was shown to be very effective in identifying high-performing regions.

\begin{itemize}
    \item For HER and ORR, the objective was to maximize similarity to \textit{conductivity} while minimizing similarity to \textit{dielectric}.
    \item For oxygen evolution OER, the objective was to maximize similarity to \textit{dielectric} while minimizing similarity to \textit{conductivity}.
\end{itemize}

Compositions on the reaction-specific Pareto fronts in similarity space defined by $S_i$ constitute our predictions for high-performing materials.

\section{Results}

\subsection{Centroid Convergence Over Iterations}

\begin{figure}[h!]
    \centering
    \includegraphics[width=0.75\linewidth]{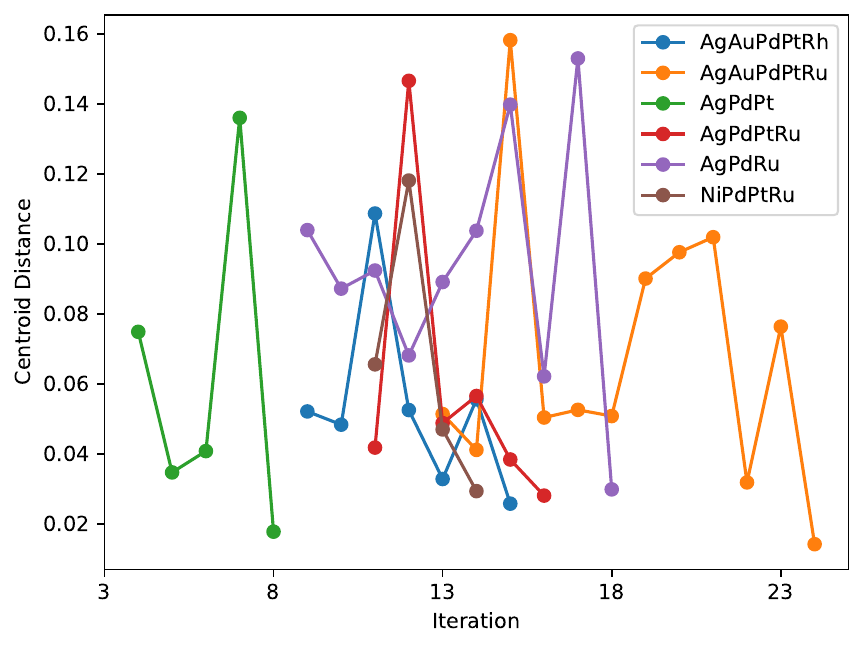}
    \caption{Centroid distance as a function of iterations for different material systems.}
    \label{fig:centroidConvergence}
\end{figure}

Figure~\ref{fig:centroidConvergence} shows the changes in centroid distance during each iteration for the six material systems we tested: AgPdPt, AgPdRu, AgPdPtRu (for ORR)~\cite{Banko2024}, AgAuPdPtRh, AgAuPdPtRu (for HER)~\cite{Thelen2025}, and NiPdPtRu (for OER)~\cite{Thelen2025a}.
The centroid distance measures the \textit{shift} in the embedding space between iterations, and serves as a quantitative indicator of convergence.
The first iteration constitutes the initialization and, therefore, does not have a previous iteration to compare against.
Additionally, not all systems have values for every iteration because the embedding space is only defined when a trained Word2Vec model provides representations for all required elements as well as the target properties \textit{dielectric} and \textit{conductivity}.
In other words, for some compositions, the first few batches of documents do not contain all tokens required to compute the word embeddings as well as their similarity scores.
If the selected corpus for a given iteration lacks such similarities, the similarity space representation can not be established and, consequently, the centroid cannot be calculated.
The number of iterations required to reach the convergence threshold of 0.03 for each material system is as follows:
\begin{itemize}
    \item \textbf{AgPdPt:} Converged after 8 iterations, indicating steady refinement and early stabilization.
    \item \textbf{AgPdRu:} Required 18 iterations to stabilize, reflecting a gradual and extended refinement process.
    \item \textbf{AgPdPtRu:} Achieved convergence in 16 iterations, suggesting moderate complexity in refining its embedding space.
    \item \textbf{AgAuPdPtRh:} Stabilized after 15 iterations, indicating consistent refinement with gradual improvement.
    \item \textbf{AgAuPdPtRu:} Required 24 iterations to reach the threshold, showing a more extended effort to refine the embedding space.
    \item \textbf{NiPdPtRu:} Converged after 14 iterations, balancing between moderate refinement complexity and stabilization.
\end{itemize}

These results for different material systems exhibit a large variability in convergence behavior.
AgPdPt has relatively fast convergence, while AgAuPdPtRu requires the largest number of iterations, reflecting the challenges of establishing a complex embedding space.
Again, once the centroid distance falls below a user-defined, heuristically obtained threshold of 0.03, we consider the embedding space converged.

\subsection{Pareto Analysis with Full Corpus vs. Selected Subset}

\begin{figure}
    \centering
    \includegraphics[width=0.75\linewidth]{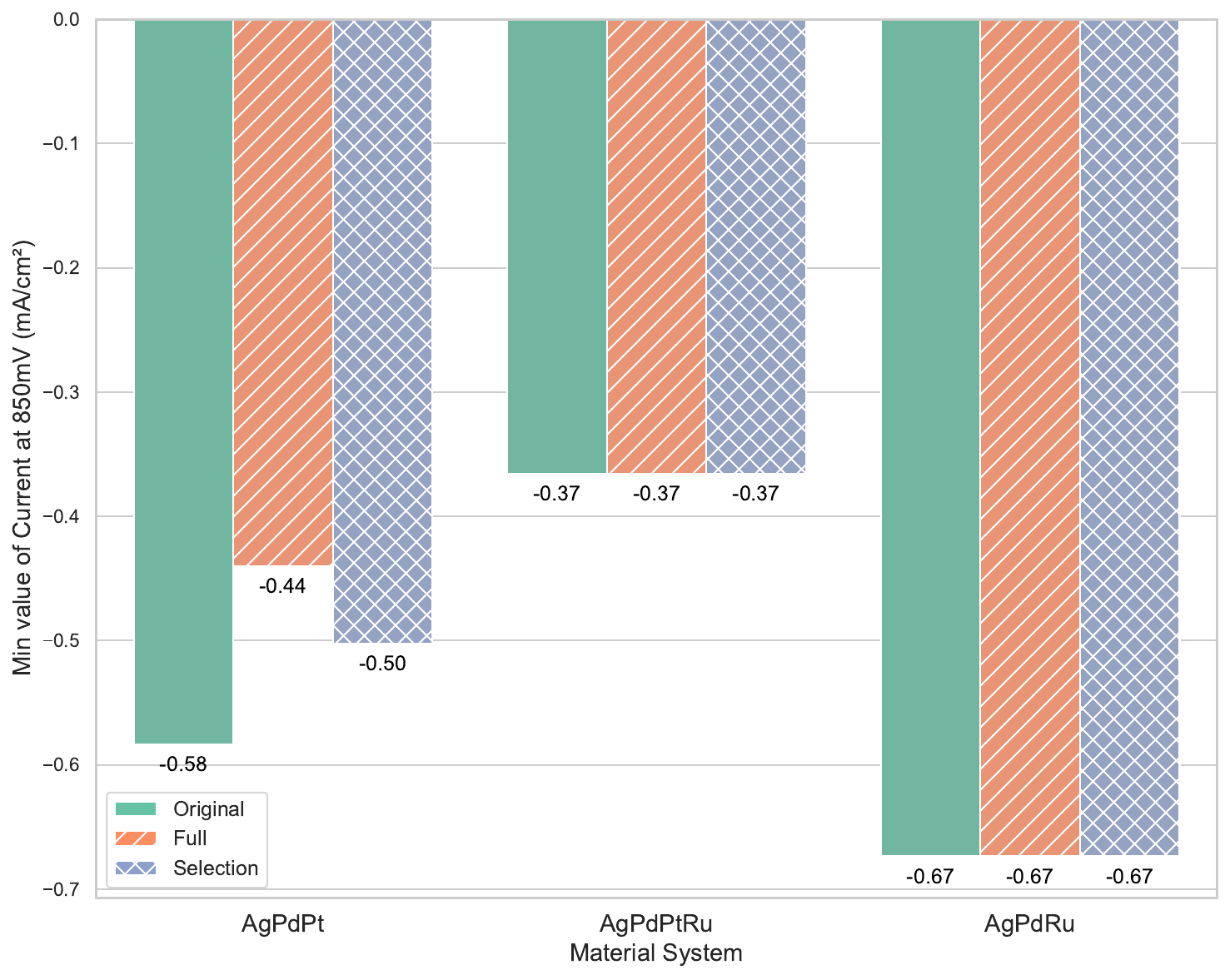}
    \caption{Comparison of minimum current density values at 850\,mV for ORR systems (AgPdPt, AgPdRu, AgPdPtRu) derived from the original performance data, the full-corpus model, and the selected-corpus model.}
    \label{fig:paretoComparisonORR}
\end{figure}

\begin{figure}
    \centering
    \includegraphics[width=0.75\linewidth]{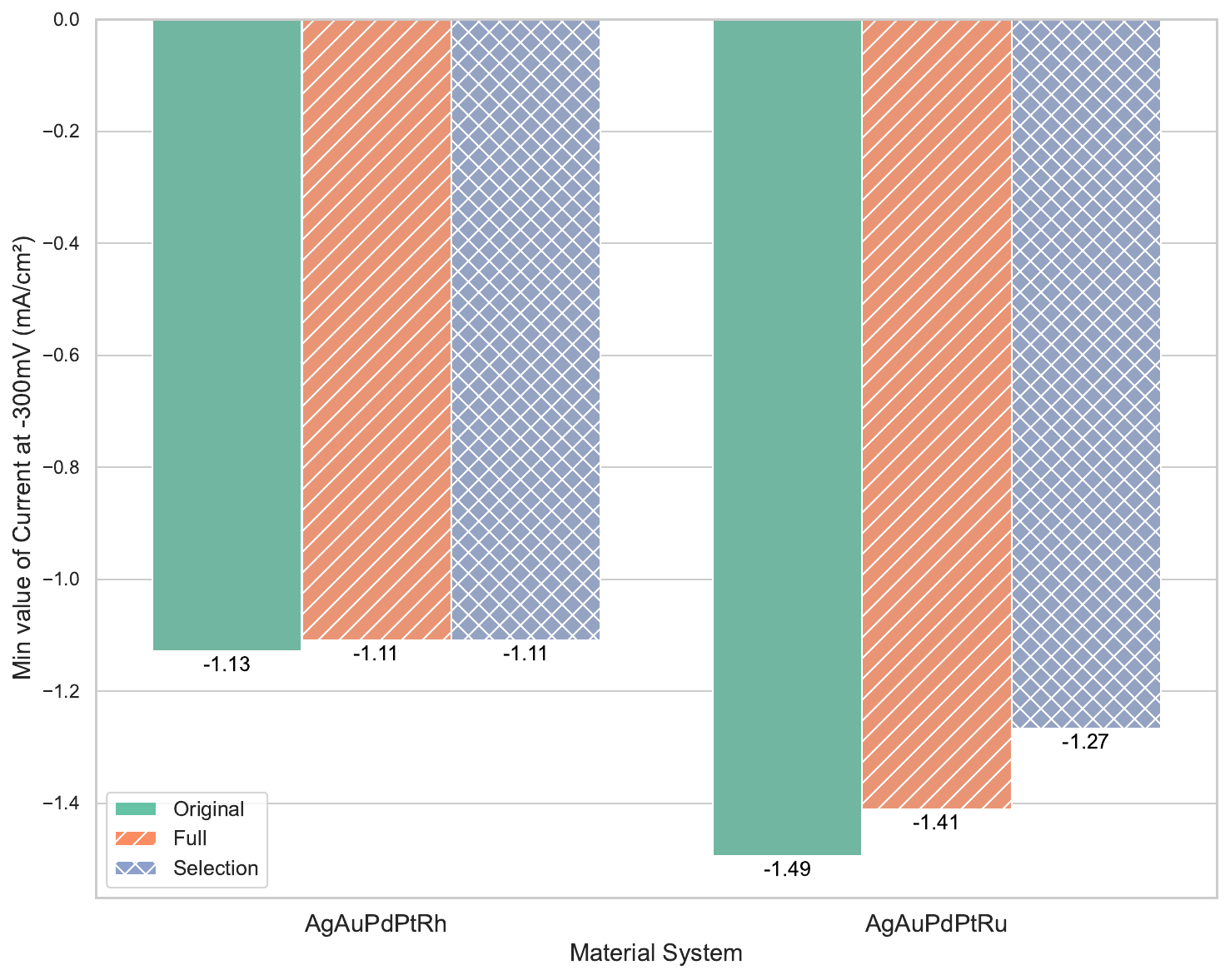}
    \caption{Comparison of minimum current density values at -300\,mV for HER systems (AgAuPdPtRh, AgAuPdPtRu) derived from the original performance data, the full-corpus model, and the selected-corpus model.}
    \label{fig:paretoComparisonHER}
\end{figure}

\begin{figure}
    \centering
    \includegraphics[width=0.75\linewidth]{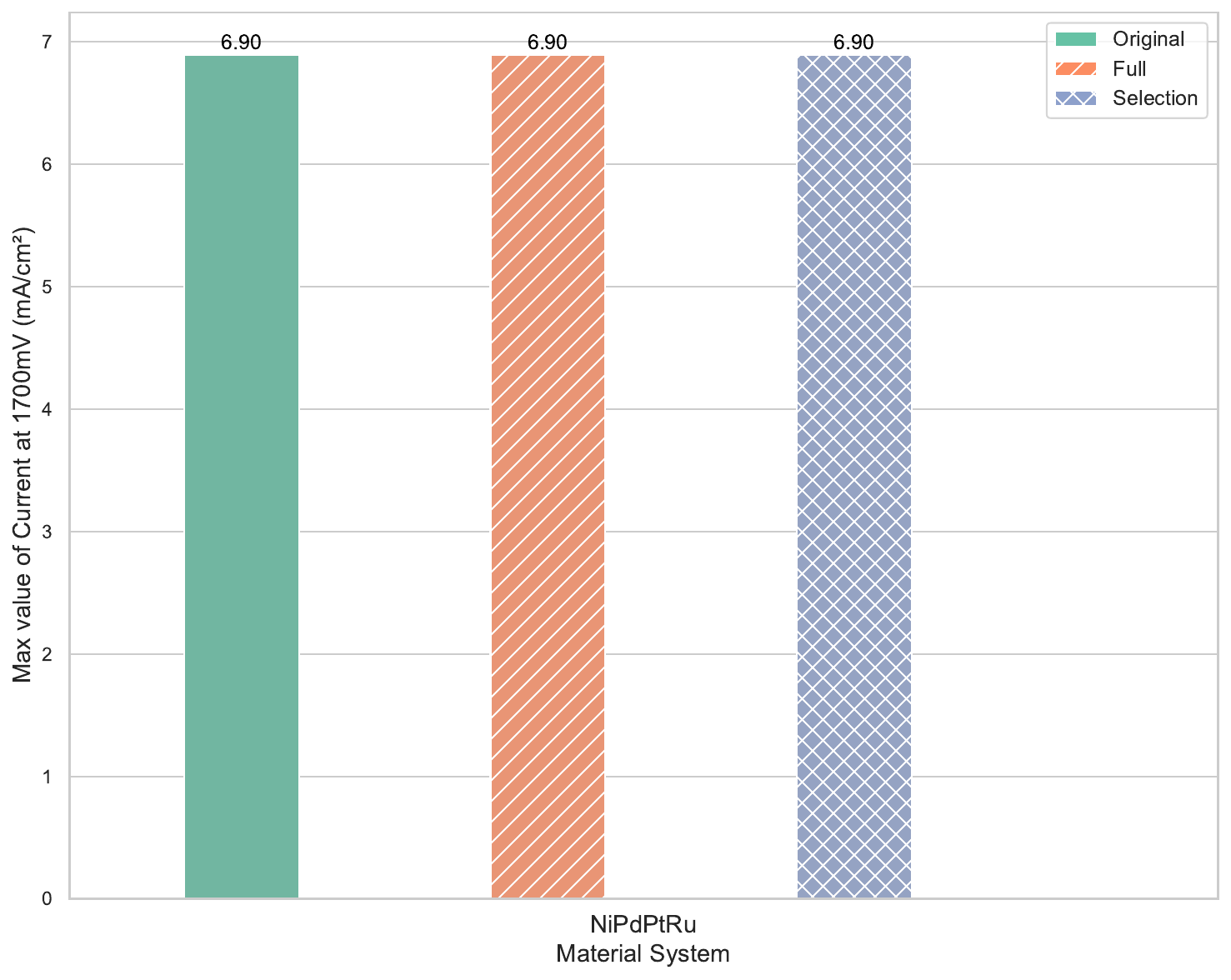}
    \caption{Comparison of maximum current density values at 1700\,mV for the OER system (NiPdPtRu) derived from the original performance data, the full-corpus model, and the selected-corpus model.}
    \label{fig:paretoComparisonOER}
\end{figure}

After centroid convergence, we perform Pareto analysis following the method presented in~\cite{Zhang2025a} on each reaction type (ORR, HER, OER) using the final Word2Vec model for each material system.
We compare the prediction of the candidate materials of the converged representations with predictions based on a model trained on all documents.
This reference full-corpus model is the same for all three materials systems.
Figures~\ref{fig:paretoComparisonORR}, \ref{fig:paretoComparisonHER}, and \ref{fig:paretoComparisonOER} summarize the prediction metrics, highlighting the models' ability to predict high-performing materials in very different material systems and reaction types.

For ORR (Fig.~\ref{fig:paretoComparisonORR}), the selected-corpus model identified high-performing materials with subsequently measured minimum current densities at 850\,mV of -0.50, -0.67, and -0.37\,mA/cm\textsuperscript{2} for the AgPdPt, AgPdRu, and AgPdPtRu systems, respectively.
These predictions align closely with best-performing experimentally measured electrocatalytic response, which recorded -0.58, -0.67, and -0.37\,mA/cm\textsuperscript{2}, respectively.
The full-corpus model, while also successful in identifying high-performing materials, predicted slightly less optimal values of -0.44, -0.67, and -0.37\,mA/cm\textsuperscript{2}.
This difference demonstrates the effectiveness of the selected-corpus model in refining the search space and focusing on the most relevant documents for a given composition space.

Fig.~\ref{fig:paretoComparisonHER} shows the performance for the AgAuPdPtRh and AgAuPdPtRu systems for HER.
The selected-corpus model achieved minimum current densities at -300\,mV of -1.13 and -1.44\,mA/cm\textsuperscript{2}, closely matching the best measured electrocatalytic response of -1.13 and -1.49\,mA/cm\textsuperscript{2}.
The full-corpus model predicts values of -1.11 and -1.41\,mA/cm\textsuperscript{2}, which, while accurate, do not represent the actual highest-performing compositions.
These results further validate the ability of the selected-corpus model to identify optimal candidates effectively.

For HER, Figure~\ref{fig:paretoComparisonHER} presents the performance for the AgAuPdPtRh and AgAuPdPtRu systems.
The selected-corpus model predicts minimum current densities at -300\,mV of -1.11 and -1.27\,mA/cm\textsuperscript{2}.
For the AgAuPdPtRu system, the value is slightly less optimal than the full-corpus model predictions of -1.41\,mA/cm\textsuperscript{2}.
Despite this, the selected-corpus model captures essential information from the full corpus, demonstrating its ability to approximate high-performing materials with a significantly smaller number of documents.

For OER, as shown in Figure~\ref{fig:paretoComparisonOER}, the maximum current density at 1700\,mV for the NiPdPtRu system is identical across all models, including the selected-corpus model, the full-corpus model, and the best measured data performance: 6.90\,\allowbreak mA/cm\textsuperscript{2}.
This, again, confirms the consistency of our approach in capturing the best-performing materials using much fewer documents for creation of the representations.

All results collectively demonstrate that both the full-corpus and selected-corpus models successfully predict high-performing materials across all reaction types.
Notably, the selected-corpus models achieve this with a much smaller number of documents.
By refining the training corpus to only the most diverse documents measured by a combination of Doc2Vec embeddings, greedy selection and convergence threshold based on the centroid change, the iterative selection process improves the predictive power of the model without compromising performance across very different scenarios.

\subsection{Current Density Measurements and Statistical Overview}

\begin{table*}[h!]
\centering
\caption{Statistical details for electrocatalysts under specified potentials. 
Abbreviations: Entries (Ori), the original number of possible compositions before text-mining selection; Entries (Full), number of compositions on the Pareto front for the Word2Vec model based on the full corpus; Entries (Selection), number of compositions on the Pareto front based on the converged model with iterative greedy selection; Selected Documents, the total number of documents used in the final iteration to train the selected-corpus model.}

\label{tbl:generalMetrics}
\small

\begin{tabular*}{\textwidth}{@{\extracolsep{\fill}}lccccc}
\hline
\begin{tabular}[c]{@{}c@{}} \textbf{Material} \\ \textbf{Systems} \end{tabular} & 
\begin{tabular}[c]{@{}c@{}} \textbf{Potential} \\ \textbf{(mV)} \end{tabular} & 
\begin{tabular}[c]{@{}c@{}} \textbf{Entries} \\ \textbf{(Ori)} \end{tabular} & 
\begin{tabular}[c]{@{}c@{}} \textbf{Entries} \\ \textbf{(Full)} \end{tabular} & 
\begin{tabular}[c]{@{}c@{}} \textbf{Entries} \\ \textbf{(Selection)} \end{tabular} & 
\begin{tabular}[c]{@{}c@{}} \textbf{Selected} \\ \textbf{Documents} \end{tabular} \\
\hline
AgPdPt      & 850  & 341  & 11  & 63  & 400  \\
AgPdRu      & 850  & 342  & 15  & 10  & 900  \\
AgPdPtRu    & 850  & 341  & 27  & 4   & 800  \\
AgAuPdPtRh  & -300 & 327  & 16  & 29  & 750  \\
AgAuPdPtRu  & -300 & 335  & 23  & 3   & 1200 \\
NiPdPtRu    & 1700 & 4026 & 374 & 168 & 700  \\
\hline
\end{tabular*}
\end{table*}

\begin{table*}[h!]
\centering
\caption{Minimum and maximum current densities (mA/cm\textsuperscript{2}) for electrocatalysts under specified potentials. 
Abbreviations: Min/Max (Ori, Full, Selection), the minimum and maximum current densities for each of the three scenarios (original data, full corpus model, selected-corpus models).}

\label{tbl:minMaxMetrics}
\small

\begin{tabular*}{\textwidth}{@{\extracolsep{\fill}}lcccccc}
\hline
\begin{tabular}[c]{@{}c@{}} \textbf{Material} \\ \textbf{Systems} \end{tabular} & 
\begin{tabular}[c]{@{}c@{}} \textbf{Min} \\ \textbf{(Ori)} \end{tabular} & 
\begin{tabular}[c]{@{}c@{}} \textbf{Min} \\ \textbf{(Full)} \end{tabular} & 
\begin{tabular}[c]{@{}c@{}} \textbf{Min} \\ \textbf{(Selection)} \end{tabular} & 
\begin{tabular}[c]{@{}c@{}} \textbf{Max} \\ \textbf{(Ori)} \end{tabular} & 
\begin{tabular}[c]{@{}c@{}} \textbf{Max} \\ \textbf{(Full)} \end{tabular} & 
\begin{tabular}[c]{@{}c@{}} \textbf{Max} \\ \textbf{(Selection)} \end{tabular} \\
\hline
AgPdPt      & -0.58 & -0.44 & -0.50 & -0.06 & -0.16 & -0.06 \\
AgPdRu      & -0.67 & -0.67 & -0.67 & -0.07 & -0.36 & -0.36 \\
AgPdPtRu    & -0.37 & -0.37 & -0.37 & -0.06 & -0.06 & -0.34 \\
AgAuPdPtRh  & -1.13 & -1.11 & -1.11 & -0.69 & -0.73 & -0.72 \\
AgAuPdPtRu  & -1.49 & -1.41 & -1.27 & -0.79 & -1.14 & -1.06 \\
NiPdPtRu    & 0.24  & 0.60  & 0.53  & 6.90  & 6.90  & 6.90  \\
\hline
\end{tabular*}
\end{table*}

Table~\ref{tbl:generalMetrics} and Table~\ref{tbl:minMaxMetrics} present the statistical details and measured current densities (in mA/cm\textsuperscript{2}) for each system at its corresponding potential. 

\noindent \textbf{ORR Systems (AgPdPt, AgPdRu, AgPdPtRu at 850~mV):}
\begin{itemize}
    \item In \textbf{AgPdPt}, the selection-based model retains 63 entries compared to 11 in the full-corpus model. The minimum current density shifts from $-0.44$\allowbreak mA/cm\textsuperscript{2} in the full model to $-0.50$~mA/cm\textsuperscript{2} in the selection model, while the maximum rises to $-0.06$~mA/cm\textsuperscript{2}.
    \item In \textbf{AgPdRu}, the minimum and maximum current densities remain the same across selection-based model and full-corpus model, at $-0.67$~mA/cm\textsuperscript{2} and $-0.36$~mA/cm\textsuperscript{2}, respectively, with 10 entries retained in the selection model compared to 15 in the full model.
    \item In \textbf{AgPdPtRu}, the selection-based model retains 4 entries compared to 27 in the full model. The minimum current density remains $-0.37$~mA/cm\textsuperscript{2}, while the maximum is lower at $-0.34$~mA/cm\textsuperscript{2} compared to $-0.06$~mA/cm\textsuperscript{2} in the full model.
\end{itemize}

\noindent \textbf{HER Systems (AgAuPdPtRh, AgAuPdPtRu at $-300$~mV):}
\begin{itemize}
    \item In \textbf{AgAuPdPtRh}, the selection-based model retains 29 entries compared to 16 in the full model. The minimum current density is consistent at $-1.11$~\allowbreak mA/cm\textsuperscript{2}, while the maximum slightly increases from $-0.72$~mA/cm\textsuperscript{2} in the selection model to $-0.73$~mA/cm\textsuperscript{2} in the full model.
    \item In \textbf{AgAuPdPtRu}, the selection-based model retains 3 entries compared to 23 in the full model. The minimum current density slightly increases from $-1.41$~mA/cm\textsuperscript{2} in the full model to $-1.27$~mA/cm\textsuperscript{2} in the selection model, while the maximum decreases from $-1.14$~mA/cm\textsuperscript{2} to $-1.06$~mA/cm\textsuperscript{2}.
\end{itemize}

\noindent \textbf{OER System (NiPdPtRu at 1700~mV):}
\begin{itemize}
    \item \textbf{NiPdPtRu} starts with a large set of 4026 candidate compositions. The selection-based model retains 168 entries compared to 374 in the full-corpus model. The minimum current density decreases from $0.60$~mA/cm\textsuperscript{2} in the full model to $0.53$~mA/cm\textsuperscript{2} in the selection model, while the maximum remains consistent at $6.90$~mA/cm\textsuperscript{2}.
\end{itemize}

All these results demonstrate that the iterative selection-based approach effectively narrows the dataset while preserving or even improving the ability to predict high-performing electrocatalysts from a large list of possible candidate compositions.
The evidence in Table~\ref{tbl:minMaxMetrics} supports the conclusion that a focused corpus can adequately capture and even surpass predicting the key characteristics of the full-corpus dataset.

\section{Discussion}

Our iterative selection approach consistently results in models that match or exceed the performance of the model built using the full corpus.
For ORR and HER systems, the method identifies more negative (better) current densities, indicating higher catalytic performance.
For OER, the selected-corpus model suggests the same high-performing compositions as the full corpus model.
These results confirm that reducing the training set---when done strategically---does not compromise, and even improves the capability to predict high-performance compositions.

\subsection{Advantages of Iterative Corpus Selection}
A feature of our approach is the \emph{stability criterion} that checks whether adding more information significantly changes the model’s representation of materials.
Once the difference between subsequent centroids in embedding space stays below a user-defined, heuristically found threshold, we stop adding more information.
Note, all models use the same threshold. This ensures:
\begin{enumerate}
    \item \textbf{Reduced Noise:} Unnecessary or very similar information is left out, preventing additional fuzziness of key semantic signals related to electrocatalytic properties.
    \item \textbf{Computational Efficiency:} Fewer documents result in faster training and analysis.
    \item \textbf{Focus:} The final Word2Vec embeddings capture terminology tailored to the studied compositions, reactions, and the properties \textit{dielectric} and \textit{conductivity}.
\end{enumerate}

\subsection{Practical Impact on Electrocatalyst Screening}
By identifying fewer yet relevant documents for training, this approach can speed up the overall pipeline of electrocatalyst design in particular for scenarios where experimental data are scarce and simulation-based data prove computationally intractable.
Fewer experimental tests are needed because the Pareto analysis, driven by these tailored embeddings, allows to narrow down a large pool of candidate compositions to a much small, experimentally accessible likely high-performing compositions.

\subsection{Mitigating the Growing Volume of AI-Generated Text}
The rise of AI-generated information, specifically texts, poses new challenges.
Large language models sometimes produce repetitive, paraphrased, or even wrong content by hallucination~\cite{Zhang2023a}.
On a large scale, if generative AI models which are retrained on recursively generated output (meaning their own output), the resulting models become defective -- they start producing noise~\cite{Shumailov2024}.
Our iterative selection loop, which operates in the embedding space and discards information that does not shift the centroid meaningfully, offers a possibility to filter noise or at least avoid duplicated information: our framework avoids documents which contribute little new information.
As AI-generated content continues to grow, such filtering mechanisms may become essential to maintaining reliable text-based models.

\subsection{Limitations and Potential Extensions}

While the proposed framework demonstrates significant strengths in refining training corpora and often even improves predictive performance, it is not without limitations. A critical challenge lies in balancing coverage and specificity.
If the initial subset of documents is too small or biased, it risks omitting important details, potentially reducing the general applicability of the model.
Additionally, reliance on predefined target properties, such as \textit{dielectric} and \textit{conductivity} as in our case require domain knowledge to be defined.
Such terms might be difficult to find for other prediction scenarios and broader applicability of our method.
Future work will address these challenges by exploring the following extensions:
\begin{itemize}
    \item \textbf{Human in the loop:} Involve domain experts to review and validate the selected documents, ensuring that important topics or underrepresented domains are not overlooked. For example, experts could identify emerging areas or validate key materials that align with research objectives.
    \item \textbf{Hybrid Models:} Integrate advanced NLP methods, such as transformer-based encoders (e.g., MatBERT~\cite{Trewartha2022}), to complement the iterative selection process. These models may provide additional layers of semantic evaluation or help identify nuanced relationships that Word2Vec with fixed representations might miss.
    \item \textbf{Property Expansion:} Extend the framework to incorporate a broader range of material properties or performance metrics. For instance, multi-objective optimization could include properties like catalytic efficiency, stability under reaction conditions, or environmental impact, enabling more targeted materials discovery and optimization pipelines.
\end{itemize}

\section{Conclusion}

Our contribution introduces an iterative framework for refining scientific corpora to create word2vec representations of with a minimal set of documents for specific prediction scenarios.
By dynamically selecting a subset of `most diverse' documents sampled in Doc2Vec embedding space with a greedy selection strategy and subsequently training Word2Vec models, our approach results in representations with minimal noise w.r.t. user-defined composition spaces and target properties as well as their relationships.
Applied to material systems for ORR, HER, and OER, our framework reliably identifies high-performing compositions from a large candidate pool using fewer training documents, matching or exceeding the performance of models trained on full corpora.
Our predictions are verified with experimental data.
This verification demonstrates the ability of our framework to efficiently use text-based knowledge for composition-property correlations.

\begin{acknowledgments}
The authors gratefully acknowledge the financial support provided by the China Scholarship Council (CSC, CSC number: 202208360048) and funding by the Deutsche Forschungsgemeinschaft (DFG, German Research Foundation) through CRC 1625, project number 506711657, subprojects INF, A05.

\end{acknowledgments}

\bibliography{apssamp}

\end{document}